\documentclass[conference,onecolumn]{IEEEtran}
\IEEEoverridecommandlockouts

\usepackage{cite}
\usepackage{amsmath,amssymb,amsfonts}
\usepackage{algorithmic}
\usepackage{graphicx}
\usepackage{textcomp}
\usepackage{xcolor}
\usepackage{soul,color}
\usepackage{subcaption}
\usepackage{comment}
\usepackage{hyperref}

\def\BibTeX{{\rm B\kern-.05em{\sc i\kern-.025em b}\kern-.08em
    T\kern-.1667em\lower.7ex\hbox{E}\kern-.125emX}}

\newif\ifanon
\anonfalse  

\begin{document}

\title{Variational Autoencoders-based Detection of Extremes in Plant Productivity in an Earth System Model}
\ifanon
\else 
\author{\IEEEauthorblockN{1\textsuperscript{st} Bharat Sharma$^*$}
\IEEEauthorblockA{\textit{Computational Engineering and Sciences Division} \\
\textit{Oak Ridge National Laboratory}\\
Oak Ridge, TN, USA \\
orcid.org/0000-0002-6698-2487}
\and
\IEEEauthorblockN{1\textsuperscript{st} Jitendra Kumar$^*$}\thanks{\textit{$^*$Bharat Sharma and Jitendra Kumar are co-first authors.}}
\IEEEauthorblockA{\textit{Environmental Sciences Division} \\
\textit{Oak Ridge National Laboratory}\\
Oak Ridge, TN, USA \\
orcid.org/0000-0002-0159-0546}
}
\fi
\maketitle

\begin{abstract}
Climate anomalies significantly impact terrestrial carbon cycle dynamics, necessitating robust methods for detecting and analyzing anomalous behavior in plant productivity. 
This study presents a novel application of variational autoencoders (VAE) for identifying extreme events in gross primary productivity (GPP) from Community Earth System Model version 2 simulations across four AR6 regions in the Continental United States. We compare VAE-based anomaly detection with traditional singular spectral analysis (SSA) methods across three time periods: 1850--80, 1950--80, and 2050--80 under the SSP5-8.5 scenario. The VAE architecture employs three dense layers and a latent space with an input sequence length of 12 months, trained on a normalized GPP time series to reconstruct the GPP and identifying anomalies based on reconstruction errors. Extreme events are defined using 5\textsuperscript{th} percentile thresholds applied to both VAE and SSA anomalies. Results demonstrate strong regional agreement between VAE and SSA methods in spatial patterns of extreme event frequencies, despite VAE producing higher threshold values (179--756 GgC for VAE vs. 100--784 GgC for SSA across regions and periods). Both methods reveal increasing magnitudes and frequencies of negative carbon cycle extremes toward 2050--80, particularly in Western and Central North America. The VAE approach shows comparable performance to established SSA techniques, while offering computational advantages and enhanced capability for capturing non-linear temporal dependencies in carbon cycle variability. Unlike SSA, the VAE method does not require one to define the periodicity of the signals in the data; it discovers them from the data. This research demonstrates the potential of deep learning approaches for extremes detection and provides a foundation for improved understanding of future carbon cycle risks under future conditions.

\end{abstract}


\section{Introduction}
The terrestrial biosphere plays a crucial role in the global carbon cycle, absorbing approximately 30\% of anthropogenic CO$_2$ emissions through photosynthetic processes~\cite{Friedlingstein2025_GlobalCarbonBudget, Sharma_2023_Biogeosciences}. Gross Primary Productivity (GPP), representing the total carbon uptake by terrestrial ecosystems, is a fundamental indicator of ecosystem functioning and carbon sequestration capacity~\cite{copernicus2024, Massoud2025}. However, the stability of this terrestrial carbon sink is threatened by increasingly frequent and intense meteorological extremes and ecological disturbances, which can disrupt normal patterns of terrestrial carbon cycle~\cite{Frank2015}.

The detection and characterization of extreme events in carbon cycle processes present significant methodological challenges due to the complex, non-linear nature of vegetation-climate interactions~\cite{liu2024knowledge}. Traditional approaches have relied on statistical methods such as percentile-based thresholds and singular spectral analysis (SSA) to identify anomalous behavior in biogeochemical time series~\cite{Sharma_2022_JGRB,sharma_icdm}. While these methods have proven effective, they rely on domain knowledge to predetermine temporal modes of variability and trends to account for, and thus may not fully capture the intricate temporal dependencies and non-linear patterns inherent in carbon cycle dynamics.

Recent advances in machine learning, particularly deep learning architectures, offer promising alternatives for anomaly detection in Earth system data~\cite{ale2024harnessingfeatureclusteringenhanced, egusphere-2025-2460, szwarcman2024quantizing
,Warner, Sharma_TRR_2025
}. 
Autoencoders and Variational autoencoders (VAE) have emerged as powerful tools for unsupervised anomaly detection, capable of learning complex data representations and identifying deviations from normal patterns~\cite{Jakubowski_2022,Vijai}. VAEs have demonstrated success in Earth science applications, including weather field synthesis and extreme event detection~\cite{oliveira2021controlling,egusphere-2025-2460,szwarcman2024quantizing}, and wildfires~\cite{stek2024}.

Earth System Models (ESMs) such as Community Earth System Model version 2 (CESM2) provide valuable datasets for investigating long-term carbon cycle dynamics under different scenarios~\cite{copernicus2024,Danabasoglu,Wieder}. CESM2 incorporates sophisticated representations of terrestrial biogeochemistry through the Community Land Model version 5 (CLM5), enabling detailed simulations of GPP and other carbon fluxes~\cite{Danabasoglu, RichterCESM2}. The availability of multi-century CESM2 simulations presents an opportunity to explore novel methods for extreme event detection across different time periods and climate conditions.

The AR6 reference regions established by the Intergovernmental Panel on Climate Change provide a standardized framework for regional climate analysis~\cite{ar6,Iturbide_2020}. Within the Continental United States (CONUS), four AR6 regions, Western North America (WNA), Central North America (CNA), Eastern North America (ENA), and Northern Central America (NCA), represent distinct climate zones and biogeochemical characteristics~\cite{ar6,ipcc2022atlas}.

This study addresses the critical need for enhanced methods to detect and characterize extreme events in terrestrial carbon cycling. We introduce a VAE-based approach for identifying GPP extremes in CESM2 simulations and compare its performance with established SSA techniques across multiple time periods and regions. This research contributes to methodology for detection and analysis of spatiotemporal patterns of extremes in carbon cycle while also advancing the application of machine learning in Earth System sciences.


\section{Data and Methods}
\subsection{CESM2 Model Data}
\label{sec:data}
\begin{figure}[ht]
    \centering
    \includegraphics[width=0.475\textwidth, trim=0cm 0cm 0cm 1cm, clip]{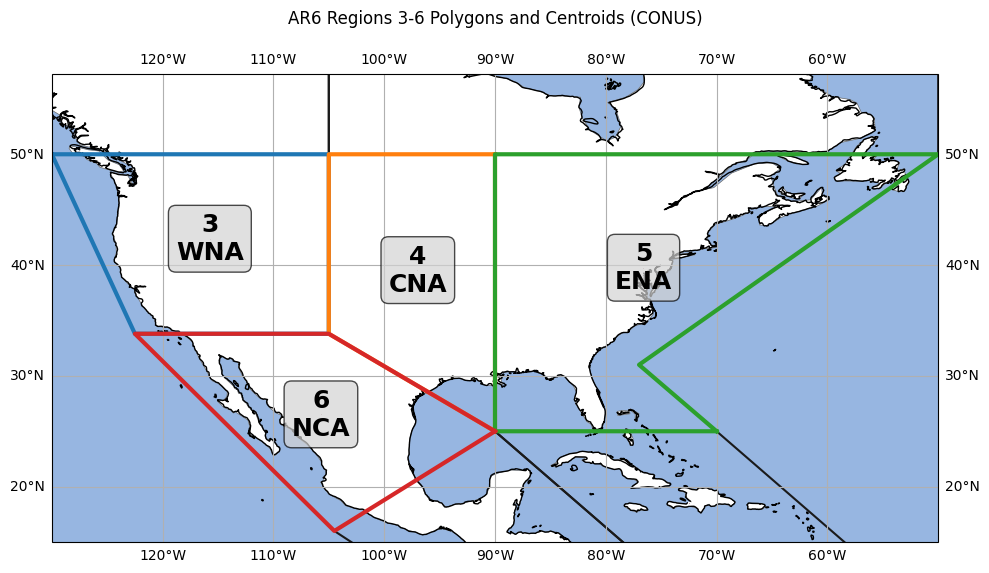} 
    \caption{US AR6 regions reference map.}
    \label{fig:US_ar6_regions}
\end{figure}

We utilized monthly GPP output from CESM2 simulations at about 1 degree spatial resolution spanning three distinct time periods: 1850--80 (historical baseline), 1950--80 (mid-20th century), and 2050--80 (future projection under SSP5-8.5 scenario)~\cite{Danabasoglu}. The CESM2 model incorporates the CLM5 with comprehensive biogeochemical processes, providing physically consistent representations of terrestrial carbon dynamics~\cite{RichterCESM2}. GPP data were analyzed for four AR6 regions within CONUS: Western North America (WNA), Central North America (CNA), Eastern North America (ENA), and Northern Central America (NCA), see Fig.~\ref{fig:US_ar6_regions}.

\subsection{Variational Autoencoder Architecture}
Autoencoders are unsupervised neural network models designed to learn compressed, low-dimensional representations of data. They consist of two main components: an encoder, which maps the input into a latent representation, and a decoder, which reconstructs the input from this representation. The network is trained to minimize the reconstruction error between the input and output, encouraging the latent space to capture informative features of the data. \cite{kingma2013vae} proposed variational autoencoders that extended autoencoders by introducing a continuous and probabilistic latent space representation, unlike discrete and fixed representation of autoencoders. 
The VAE encoder maps the input variables to mean ($\mu$) and log variance ($\log\sigma^2$) (Eq.~\ref{eq:mu-sigma}), defining the distribution in the latent space ($z \sim \mathcal{N}(0, I)$).

\begin{equation}
\mu = \mathrm{Encoder}_\mu(X),\quad \log\sigma^2 = \mathrm{Encoder}_\sigma(X)
\label{eq:mu-sigma}    
\end{equation}

\noindent
However, direct sampling from the latent space distribution ($\mathcal{N}(0, I)$) violates the differentiable property desired for gradient descent backpropagation, since vector of random values have no derivative. Thus, \cite{kingma2013vae} introduced a reparameterization trick by introducing a new parameter, $\epsilon$, which is a random value selected from the normal distribution between 0 and 1, reparameterizing latent variable $z$ as in Eq.~{\ref{eq:reparam_trick}. Since the random value $\epsilon$ is not derived from and has no relation to the autoencoder model’s parameters, it can be ignored during backpropagation.

\begin{equation}
z = \mu + \sigma \odot \epsilon,\quad \epsilon\sim \mathcal{N}(0, I)
\label{eq:reparam_trick}
\end{equation}

The decoder reconstructs the input (Eq.~\ref{eq:decoder}):
\begin{equation}
\hat{X} = \mathrm{Decoder}(z)    
\label{eq:decoder}
\end{equation}

\noindent
The loss function (Eq.~\ref{eq:kl_loss}) combines the reconstruction loss and Kullback-Leibler divergence (Eq.~\ref{eq:kl_divergence}) to constrain the encoder output to follow standard normal distribution $\mathcal{N}(0, I)$. 

\begin{equation}
\mathcal{L}_\mathrm{VAE} = 
\underbrace{\mathbb{E}_{q_{\phi}(z \mid x)} \big[ \log p_{\theta}(x \mid z) \big]}_{\text{Reconstruction Loss}}
\;+ \beta\;
\underbrace{D_{\mathrm{KL}} \!\left( q_{\phi}(z \mid x) \;\|\; p(z) \right)}_{\text{Regularization (KL Divergence)}}
\label{eq:kl_loss}
\end{equation}

\begin{equation}
D_{\mathrm{KL}}\!\big(q(z \mid x) \;\|\; p(z)\big) \;=\; -\tfrac{1}{2} \sum_{i=1}^{d} \Big( 1 + \log \sigma_i^2 - \mu_i^2 - \sigma_i^2 \Big)
\label{eq:kl_divergence}
\end{equation}

\noindent
where $q_{\phi}(z \mid x)$ is the encoder, $p_{\theta}(x \mid z)$ is decoder, $p(z)$ is prior distribution on the latent space, $\beta=0.5$ balances reconstruction and regularization.

We implemented VAE using the PyTorch library \cite{pytorch} and Optuna framework \cite{optuna} for hyperparameter optimization.

\subsection{Singular Spectral Analysis Implementation}

To provide baseline comparison for the results from the VAE approach, we implemented SSA following established methodologies for carbon cycle analysis~\cite{sharma2022analysis}. SSA decomposes time series into independent components representing different periodicities, enabling separation of trend (periodicities of 10 years and more) and annual cycle (12 months and its harmonic frequencies)~\cite{Sharma_Codes_Zenodo}. 
The method effectively captures non-linear trends and modulated annual cycles, making it particularly suitable for biogeochemical time series analysis under changing climate conditions.
The anomalies are computed by removing the non-linear trends and modulated annual cycle from the original timeseries, capturing inter-annual variations ($>$1 and $<$10 years of periodicity) and intra-annual variations ($<$12 months). Large-scale climate variability, such as El Nino Southern Oscillation (ENSO) which has return periods between 2 and 7 years\cite{Sharma_2022_JGRB}, are known to have significant impact on climate and ecological extremes and are captured by SSA derived anomalies.

\subsection{Extreme Event Definition and Threshold Calculation}
Following protocols for extreme event identification proposed by~\cite{Sharma_2023_Biogeosciences}, we defined extreme events using 5\textsuperscript{th} percentile thresholds applied to GPP anomalies from both VAE and SSA methods. This approach ensures that 5\% of the most extreme negative and positive anomalies are classified as extreme events, providing a consistent framework for comparison across methods and time periods.
Anomalies from the VAE method were calculated as the difference between original and reconstructed GPP values at each grid cell. For SSA, anomalies represent the residual after removing trend and annual cycle components. Thresholds were computed separately for each region and time period, allowing for adaptation to changing baseline conditions and variability characteristics.

It is important to note that the length of the reconstructed GPP time series is shortened by the sequence length used in the VAE method (12 months in this case). Because the model requires a full input window to perform reconstruction, it cannot generate values at the very beginning and end of the record, leading to a reduced effective time span. To calculate extremes consistently, we therefore remove the first and last year of anomalies from both the VAE- and SSA-derived series. Consequently, results reported for the period 1850–80 actually correspond to 1851–79, and similarly for other intervals. For clarity and readability, however, we present the original periods in the text.

\subsection{Spatial and Temporal Analysis}
Regional aggregation of extreme event frequencies and magnitudes was performed by integrating results across all grid cells (with land fraction more than 10\%) within each AR6 region. Time series analysis focused on monthly patterns of extreme event occurrence and magnitude, enabling identification of seasonal and inter-annual variations. Spatial distribution patterns were analyzed using gridded frequency maps showing the occurrence of extreme events at individual grid cells.

\section{Results}

\subsection{Variational Autoencoder Model}
We implemented a VAE using Pytorch \cite{pytorch} to model the GPP timeseries. Driven by heterogeneity in vegetation distribution, GPP shows spatial and temporal variability across the CONUS. We trained separate models for four selected AR6 regions and for three time periods as described in Section~\ref{sec:data}. To capture the annual seasonality of vegetation and vegetation productivity well, we use a sequence of 12 months of GPP values as input to the VAE. Choice of 12 months as sequence length was made after testing varying sequence length from 6, 12, 18, 24 and 36 months. After conducting a number of trials with various combinations of hyperparameters using Optuna library \cite{optuna} (number of hidden layers: [1, 2, 3]; size of hidden layers: [8, 16, 32, 64, 128, 256]; size of latent dimension: [2:10] dropout rate: [0:0.2], learning rate: [1e-4:1e-2]; batch\_size: [32, 64, 128]), we selected an architecture and set of hyperparameters (Fig.~\ref{fig:vae_model}) that performed well for all regions. Our VAE model architecture consists of three dense hidden layers of sizes 128, 64 and 32 respectively, latent space dimension of 5, dropout rate of 0.01 and learning rate of 0.005. A ReLU~\cite{relu} activation function was used for each of the hidden layers and a tanh~\cite{Apicella2021} activation function as final layer of the decoder. GPP timeseries was normalized between $-$1 and 1 before input to the VAE model.

\begin{figure*}
    \centering
    \includegraphics[width=1\linewidth]{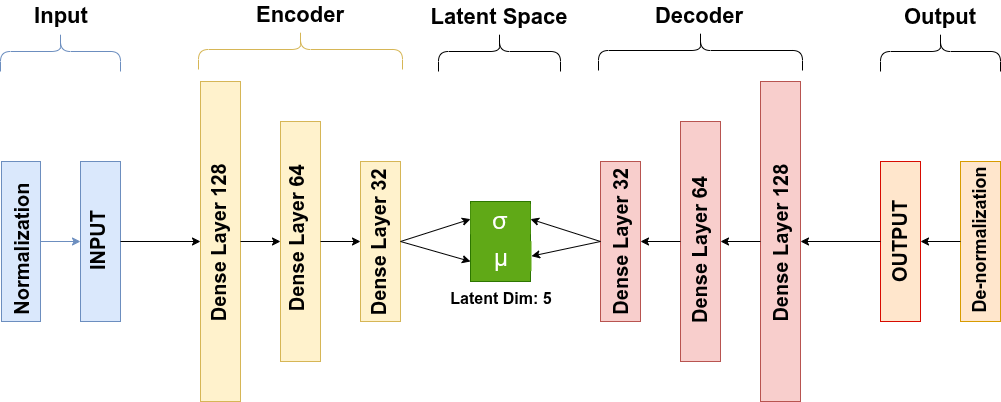}
    \caption{Architecture of variational autoencoder designed for detecting GPP extremes}
    \label{fig:vae_model}
\end{figure*}

Trained VAE model was applied to construct the GPP timeseries for all the study regions and periods. Fig.~\ref{fig:vae_orig_recon} demonstrates the effectiveness of VAE for reconstructing the GPP timeseries for all four regions for the period of 1850-1880. Time series of GPP reconstruction error (original $-$ reconstructed GPP) was further used to detect and analyze the extremes in carbon cycle.

\begin{figure}
    \centering
    \includegraphics[width=1.0\linewidth]{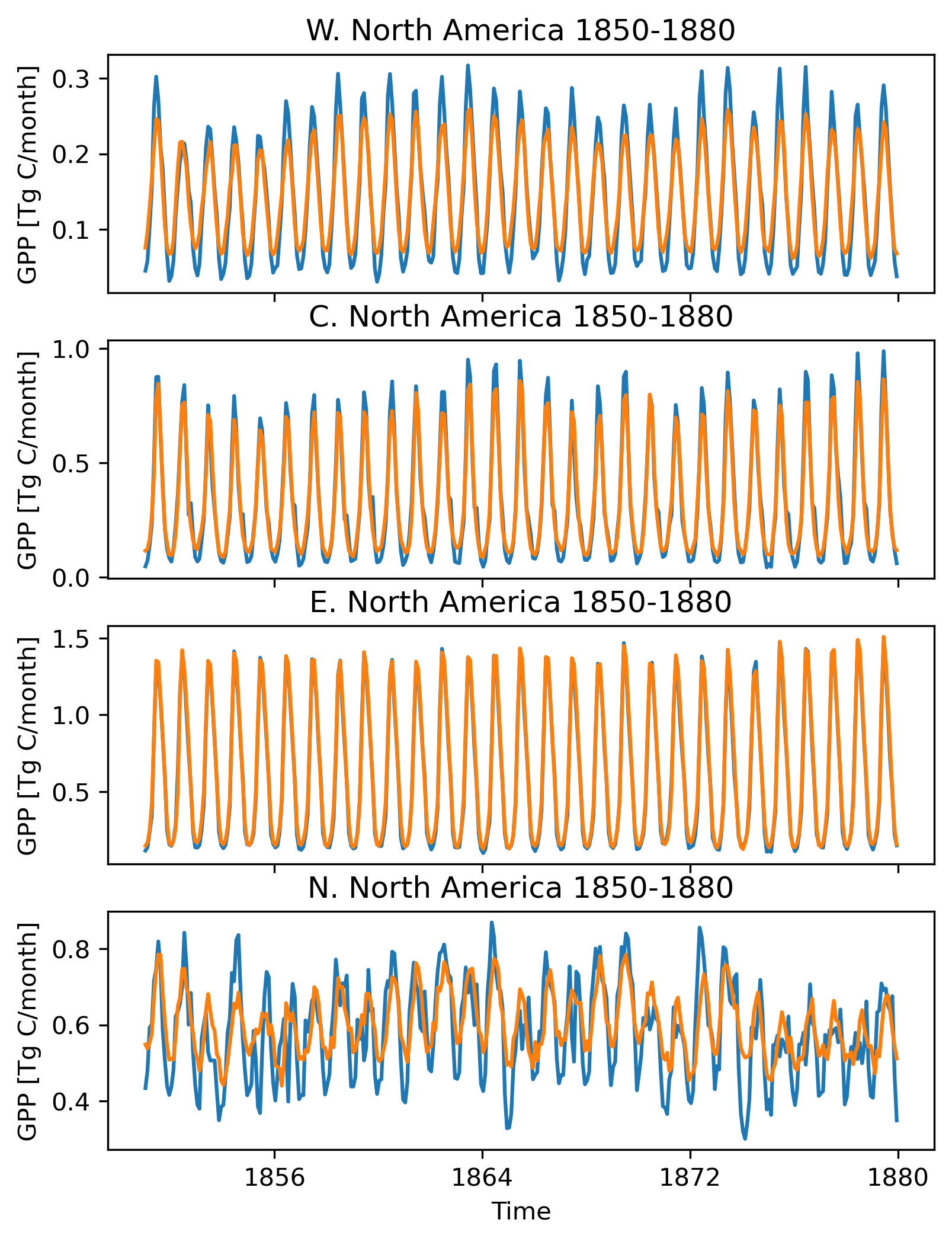}
    \caption{Comparison between original (blue line) and VAE-based reconstructed GPP time series (orange line) for the period 1850-1880 in four CONUS AR6 regions.}
    \label{fig:vae_orig_recon}
\end{figure}

Model trainings were conducted for maximum of 500 epochs. An early stopping criteria with patience of 50 was applied, stopping model training if accuracy did not improve for 50 epochs and preserving the best model. Most trainings were completed under 200 epochs. Learning rate was adaptively reduced using PyTorch \textit{ReduceLROnPlateau} with a patience of 5 and factor of 0.5. Fig.~\ref{fig:loss_curve} show the training and validation loss curves for four study regions. All model trainings were conducted on Perlmutter supercomputer at National Energy Research Scientific Computing Center (NERSC) using one AMD EPYC 7763 CPUs and four NVIDIA A100 (40~GB) GPUs.

\begin{figure}
    \centering
    \includegraphics[width=1.0\linewidth]{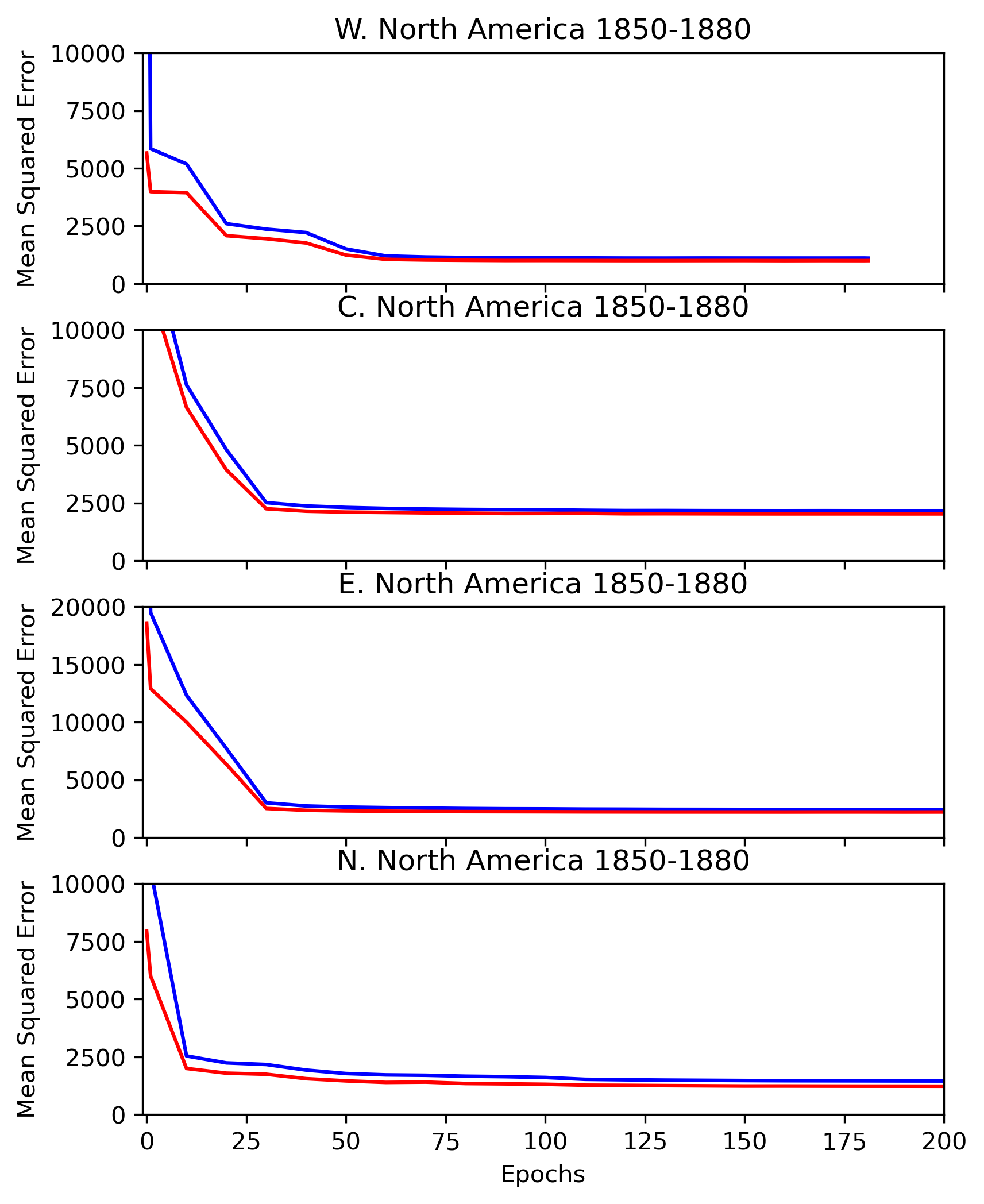}
    \caption{Convergence of model training and validation loss over epochs for the four study regions.}
    \label{fig:loss_curve}
\end{figure}

\subsection{Threshold Comparison Between VAE and SSA Methods}

The comparison of extreme event thresholds between VAE and SSA methods reveals systematic differences across regions and time periods (Table~\ref{tab:threshold}). VAE consistently produces higher threshold values than SSA, with differences varying by region and temporal period. For the WNA region, VAE thresholds range from 179 GgC (1850--80) to 457 GgC (2050--80), compared to SSA thresholds of 100--368 GgC for the same periods. This trend is consistent across most regions, with CNA and ENA showing similar patterns of higher VAE thresholds.

\begin{table}[]
\centering
\caption{Threshold for Extremes}

\label{tab:threshold}
\resizebox{\columnwidth}{!}{%
\begin{tabular}{|c|c|c|c|}
\hline
Region & \textbf{Period} & \textbf{VAE (GgC)} & \textbf{SSA (GgC)} \\ \hline
\textbf{WNA} & 1850--80 & 179 & 100 \\ \hline
\textbf{WNA} & 1950--80 & 255 & 171 \\ \hline
\textbf{WNA} & 2050--80 & 457 & 368 \\ \hline
\textbf{CNA} & 1850--80 & 302 & 321 \\ \hline
\textbf{CNA} & 1950--80 & 520 & 412 \\ \hline
\textbf{CNA} & 2050--80 & 683 & 515 \\ \hline
\textbf{ENA} & 1850--80 & 308 & 211 \\ \hline
\textbf{ENA} & 1950--80 & 401 & 263 \\ \hline
\textbf{ENA} & 2050--80 & 462 & 324 \\ \hline
\textbf{NCA} & 1850--80 & 526 & 503 \\ \hline
\textbf{NCA} & 1950--80 & 635 & 510 \\ \hline
\textbf{NCA} & 2050--80 & 756 & 784 \\ \hline
\end{tabular}%
}
\end{table}

The increasing magnitude of thresholds from 1850--80 to 2050--80 demonstrates the intensification of extreme events under projected future climate conditions across both methodologies. The NCA region exhibits the highest threshold values, most likely due to large variations in the plant productivity of tropical forests that dominate the region.

\subsection{Spatial Distribution of Extreme Events}

Spatial analysis of negative extreme events reveal distinct regional hotspots and consistent patterns between VAE and SSA methods despite differences in absolute threshold values. The frequency maps (Fig.~\ref{fig:FreqSpatial_neg_1850} and \ref{fig:spatial_neg_2050}) for 1850--80 and 2050--80 periods demonstrate remarkable agreement in identifying regions of elevated extreme event occurrence, particularly in the western portions of the WNA region (including central valley and coastal region of California, and southern Sierra mountains), central areas of the CNA region (spanning arid shrublands and grasslands) and deciduous forests along Appalachian mountain range in ENA.

\begin{figure}[]
    \centering
    \begin{subfigure}[t]{0.475\textwidth}
        \centering
        \includegraphics[width=\textwidth,trim=0cm 0cm 0cm 1cm, clip]{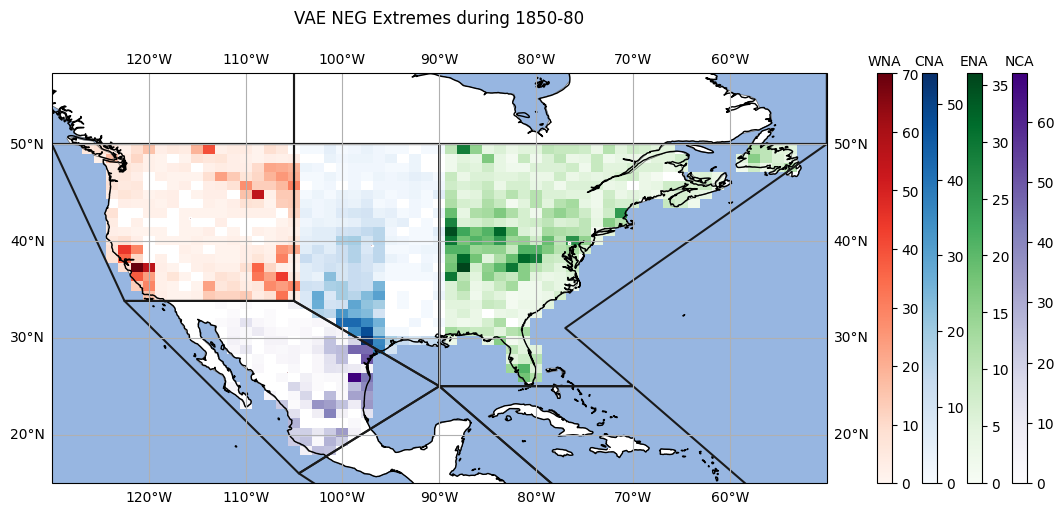}
        \caption{VAE}
        \label{fig:FreqSpatial_neg_VAE_1850}
    \end{subfigure}
    \vspace{1em}
    \begin{subfigure}[t]{0.475\textwidth}
        \centering
        \includegraphics[width=\textwidth,trim=0cm 0cm 0cm 1cm, clip]{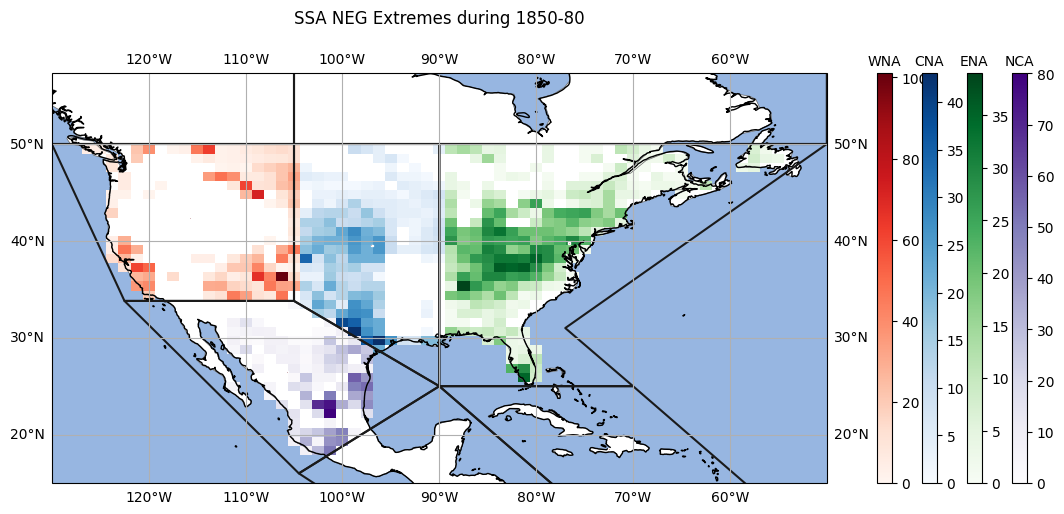}
        \caption{SSA}
        \label{fig:FreqSpatial_neg_SSA_1850}
    \end{subfigure}
    \caption{Spatial distribution of total negative extremes in GPP during 1850--1880, for VAE (top) and SSA (bottom) for each of the AR6 regions analyzed separately. A separate color schemes are used for each region to highlight the gradients.}
    \label{fig:FreqSpatial_neg_1850}
\end{figure}

For the historical period (1850--80), both VAE and SSA identify moderate frequencies of 10-30 extreme events across most regions, with localized hotspots reaching 40-50+ events in specific areas. The spatial coherence between methods suggests robust identification of ecologically vulnerable regions even with different analytical approaches.

\begin{figure}[]
    \centering
    \begin{subfigure}[t]{0.475\textwidth}
        \centering
        \includegraphics[width=\textwidth,trim=0cm 0cm 0cm 1cm, clip]{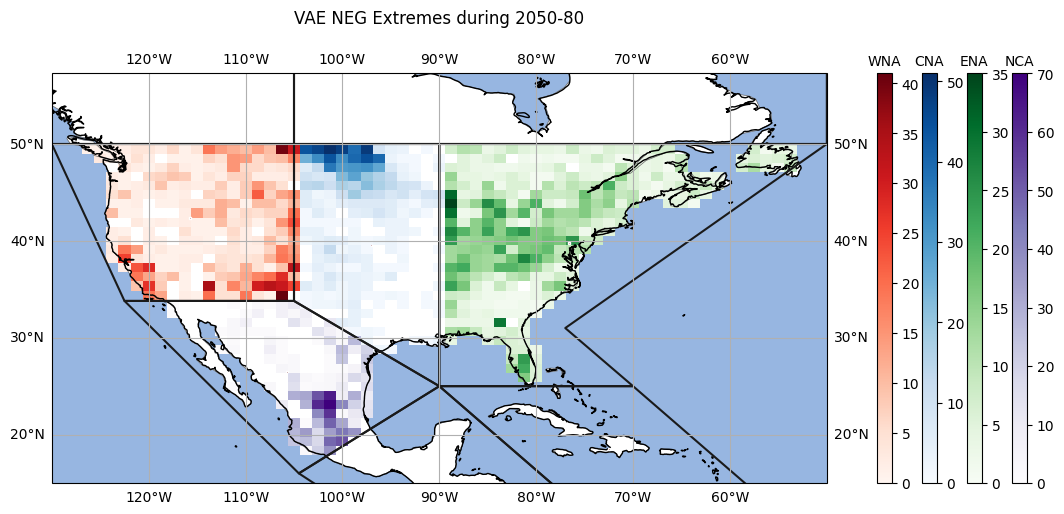}
        \caption{VAE}
    \end{subfigure}
    \vspace{1em}
    \begin{subfigure}[t]{0.475\textwidth}
        \centering
        \includegraphics[width=\textwidth,trim=0cm 0cm 0cm 1cm, clip]{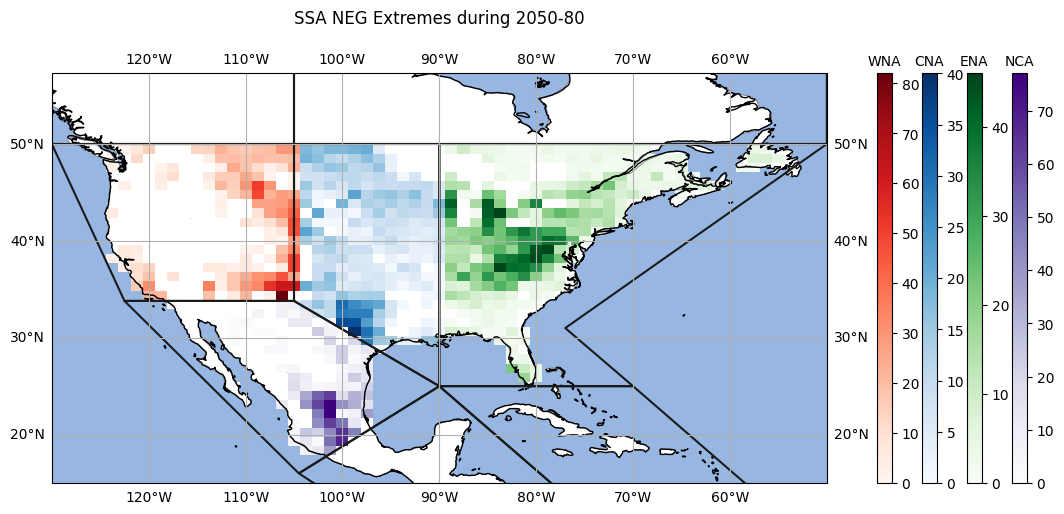}
        \caption{SSA}
    \end{subfigure}
    \caption{Spatial distribution of total negative extremes in GPP during 2050--2080, for VAE (top) and SSA (bottom) for each of the AR6 regions analyzed separately. A separate color schemes are used for each region to highlight the gradients.}
    \label{fig:spatial_neg_2050}
\end{figure}

Future projections (2050--80) show dramatic expansion of high-frequency zones across all regions. The VAE method identifies extensive areas experiencing 20-35+ extreme events, while SSA shows similar patterns with slightly different intensity distributions. Both methods consistently highlight the southwestern WNA region and central CNA region to experience the highest frequencies of extreme events, with some grid cells exceeding 50+ events over the 30-year period.
The strong spatial agreement between VAE and SSA methods, despite their different mathematical foundations, provides robust evidence for the identified hotspot regions and supports the reliability of future projections of carbon cycle extreme event distributions.

\subsection{Temporal Distribution of Frequency of Negative Extreme Events}

Analysis of time series of negative extreme events reveals substantial temporal variability and long-term trends across both VAE and SSA methods. The 1850--80 period (historical, Fig.~\ref{fig:TS_Freq_NegExt_1850}) shows relatively stable frequencies. Both methods capture similar timing of major extreme events, demonstrating consistency in temporal pattern detection.

\begin{figure}[ht]
    \centering
    \includegraphics[width=0.475\textwidth,trim=0cm 0cm 0cm 1cm, clip]{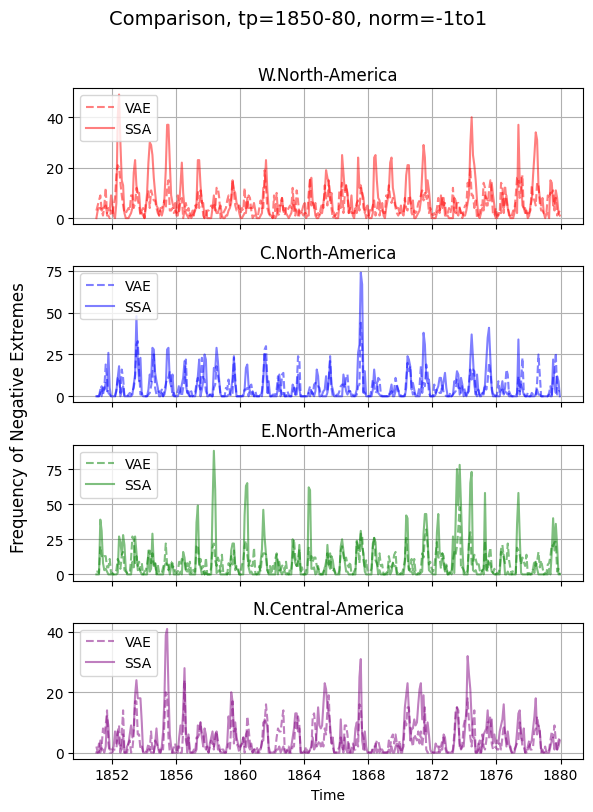}
    \caption{Time series of sum of all negative extremes in GPP for four AR6 regions during 1850--80.}
    \label{fig:TS_Freq_NegExt_1850}
\end{figure}

The 2050--80 period (Fig.~\ref{fig:TS_Freq_NegExt_2050}) exhibit markedly different characteristics, with increased baseline frequencies and more pronounced extreme events. VAE and SSA methods show strong agreement in identifying peak extreme event periods, particularly during the late 2060s and early 2070s across multiple regions. 

\begin{figure}[ht]
    \centering
    \includegraphics[width=0.475\textwidth,trim=0cm 0cm 0cm 1cm, clip]{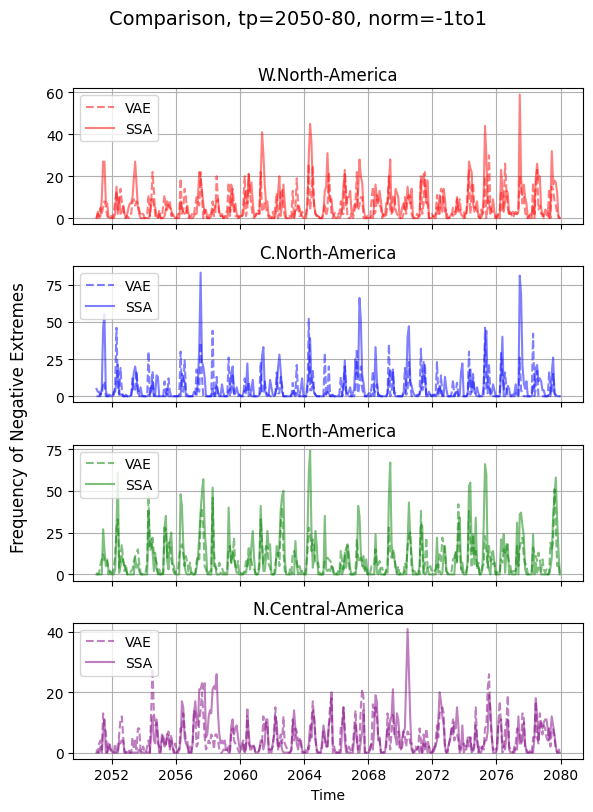}
    \caption{Time series of sum of all negative extremes in GPP for four AR6 regions during 2050--80.}
    \label{fig:TS_Freq_NegExt_2050}
\end{figure}

The CNA region displays the most dramatic increases in negative extreme event frequency, with both methods detecting severe events reaching 75+ negative extremes compared to typical values of 10-25 extremes. 

Regional differences in temporal patterns are evident across the four AR6 regions. WNA shows consistent but moderate increases in extreme event frequency, while ENA exhibits high variability with periodic intense events. The NCA region demonstrates sustained elevated frequencies throughout the 2050--80 period, suggesting persistent stress conditions under projected future climate conditions. 

We see a 50\% increase in the frequency of negative extremes in WNA between the two periods. Since the extremes are calculated using the percentile based method, increase in negative extremes also indicate decrease in positive extremes. This indicates that over time the vulnerability of forests to losses in carbon uptake could increase.

\subsection{Magnitude of Carbon Cycle Extremes}
The magnitude of negative extreme events, measured as the intensity of carbon uptake loss during extreme events, show concerning trends toward more severe impacts. Both VAE and SSA methods consistently identify periods of significant carbon cycle disruption, with magnitudes reaching $-$10 to $-$40 TgC during the most severe events across different regions and time periods.

\begin{figure}[ht]
    \centering
    \includegraphics[width=0.48\textwidth,trim=0cm 0cm 0cm 1cm, clip]{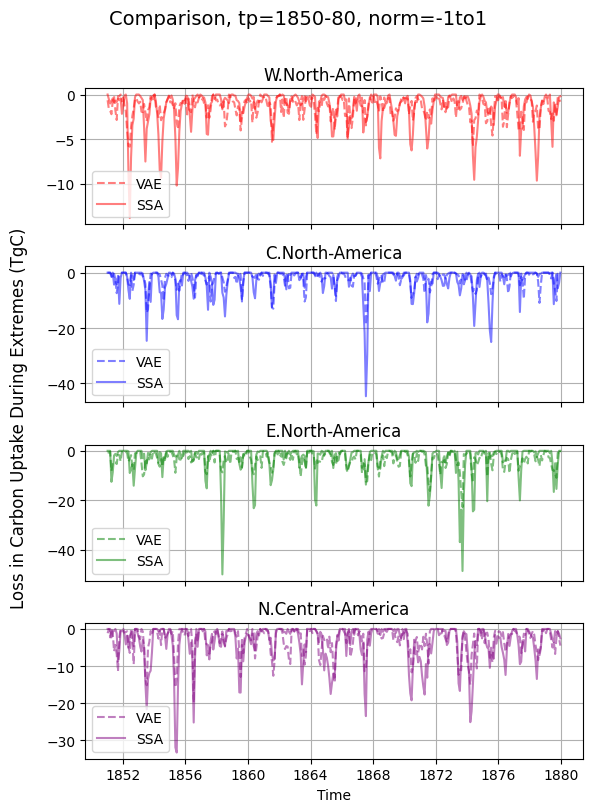}
    \caption{Time series of magnitude of sum of negative extremes in GPP in four AR6 regions during 1850--80.}
    \label{fig:TS_Mag_NegExt_1850}
\end{figure}

Historical periods (1850--80, see Fig.~\ref{fig:TS_Mag_NegExt_1850}) show relatively contained extreme event magnitudes, typically ranging from 0 to $-$10 TgC for most regions. However, future projections (2050--80, see Fig.~\ref{fig:TS_Mag_NegExt_2050}) demonstrate substantial intensification, with extreme events reaching magnitudes of $-$20 to $-$60 TgC in some regions. The CNA region exhibits the most severe projected extremes, with both VAE and SSA detecting events exceeding $-$60 TgC in magnitude during the late 2060s.

\begin{figure}[ht]
    \centering
    \includegraphics[width=0.48\textwidth, trim=0cm 0cm 0cm 1cm, clip]{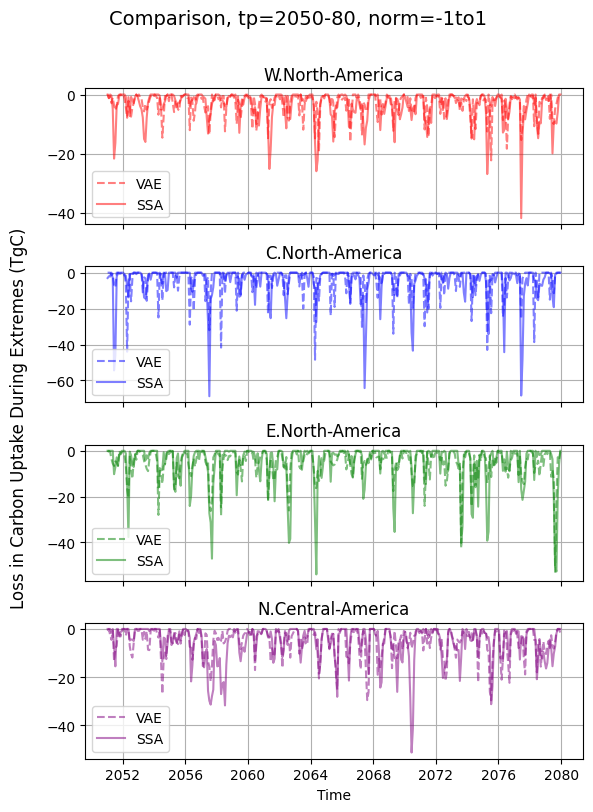}
    \caption{Time series of magnitude of sum of negative extremes in GPP in four AR6 regions during 2050--80.}
    \label{fig:TS_Mag_NegExt_2050}
\end{figure}

The consistency between VAE and SSA magnitude estimates provides confidence in the robustness of these projections. Both methods capture similar seasonal patterns in extreme event occurrence.


\section{Discussion}
The results demonstrate that VAE-based anomaly detection offers a viable alternative to traditional SSA methods for identifying extreme events in carbon cycle dynamics. The systematic differences in threshold values between VAE and SSA likely reflect the different mathematical foundations of these approaches. VAE methods learn complex non-linear patterns in data reconstruction, potentially capturing subtle anomalies that manifest as higher threshold values~\cite{ale2024harnessingfeatureclusteringenhanced, Jakubowski_2022}. In contrast, SSA relies on spectral decomposition that may be more sensitive to non-linear trend, which is calculated as sum of frequencies with periodicity of 10 years and higher, and seasonal and its harmonic components, resulting in lower threshold values for equivalent anomaly intensities~\cite{Sharma_2022_JGRB}. Unlike SSA that requires the spectral frequencies of interest to be prescribed based on expert knowledge of the processes, VAE learns the inherent patterns from the data and reconstruction errors help identify the anomalies that does not fit those learned patterns.

The strong spatial agreement between VAE and SSA methods despite threshold differences suggests both approaches identify similar underlying physical processes driving extreme events. This convergence provides confidence in the robustness of identified hotspot regions and temporal patterns. The computational advantages of VAE approaches, once trained, enable rapid processing of large climate datasets and potential real-time applications for extreme event monitoring~\cite{szwarcman2024quantizing}.

\subsection{Comparative Analysis of Positive and Negative Extreme Magnitudes}
\label{sec:dis_posneg}

Despite the systematic tendency of the VAE method to yield higher extreme thresholds (as indicated in Table~\ref{tab:threshold}), the estimated magnitude of negative extremes, defined as the aggregate loss in carbon uptake during extreme events, remained similar between VAE and SSA across all four AR6 regions in the projected 2050--80 future period. For instance, in 2050--80, the cumulative negative extremes (in TgC) for VAE and SSA, respectively, were as follows: Western North America (WNA): $-$1080 vs. $-$1288; Central North America (CNA):  $-$1322 vs. $-$1617; Eastern North America (ENA): $-$1783 vs. $-$1824; and Northern Central America (NCA): $-$1752 vs. $-$1938. This strong agreement is visually evident in the time series magnitude plots for negative extremes in Fig.~\ref{fig:TS_Mag_NegExt_1850} and~\ref{fig:TS_Mag_NegExt_2050}, where both methods track the largest episodes of carbon uptake loss with similar timing and depth, despite differences in extremes thresholds.

The intensification of extreme events from historical to future periods aligns with established understanding of climate-carbon feedbacks under warming scenarios~\cite{Sharma_2023_Biogeosciences}. The particularly severe projections for the CNA region reflect the vulnerability of continental interior regions to enhanced drought stress and temperature extremes. These findings support concerns about weakening terrestrial carbon sinks under continued environmental change~\cite{Frank2015}.

The reason for the relatively small discrepancy in negative extremes despite higher VAE thresholds can be traced to properties of the anomaly distributions. The $5^{th}$ percentile approach identifies the top 5\% of anomalies regardless of sign and can be sensitive to the distribution’s shape and skewness~\cite{Sharma_2023_Biogeosciences}. The VAE anomaly distributions are more positively skewed than those of SSA, resulting in greater calculated magnitudes for positive extremes (carbon uptake gains using extremes) in the future period, as seen in Fig.~\ref{fig:TS_Mag_PosExt_2050}. For 2050--80, positive extremes (TgC) for VAE vs. SSA were: WNA: 1731 vs. 1349; CNA: 2584 vs. 1647; ENA: 1751 vs. 1288; NCA: 1521 vs. 1765. Thus, the VAE method not only identifies more intense negative events but also projects a substantial shift toward higher magnitude positive carbon uptake extremes, particularly in WNA, CNA, and ENA.

\begin{figure}[ht]
    \centering
    \includegraphics[width=0.48\textwidth, trim=0cm 0cm 0cm 1cm, clip]{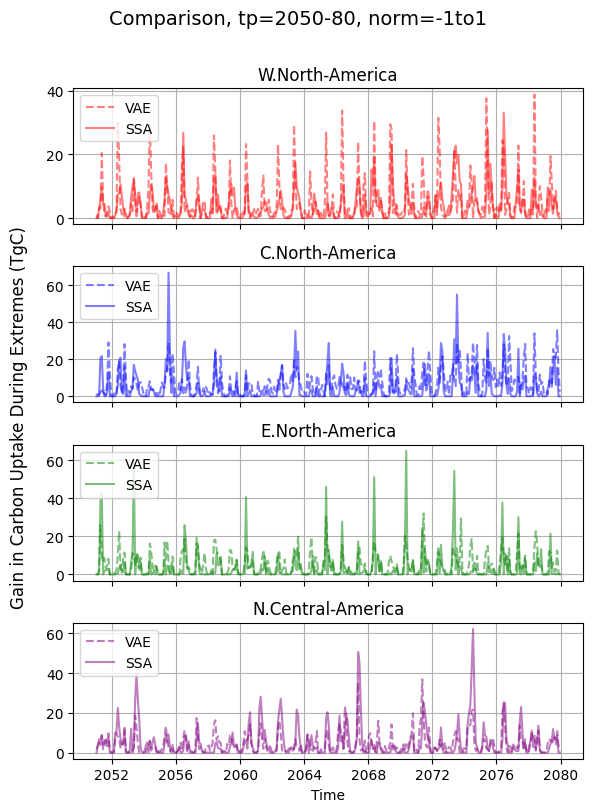}
    \caption{Time series of magnitude of sum of positive extremes in GPP in four AR6 regions during 2050--80.}
    \label{fig:TS_Mag_PosExt_2050}
\end{figure}

These findings have important implications for interpreting extremes detection under changing environmental conditions vegetation experience. The conservative (right-skewed) bias induced by VAE's latent representation causes positive extremes to be more pronounced, especially under higher atmospheric CO$_2$ concentrations with stronger seasonal carbon uptake and intermittent recovery events. Meanwhile, the negative extremes remain robust and comparable between VAE and SSA across spatiotemporal domains, suggesting that both methods identify periods of greatest carbon cycle vulnerability. This also points to the suitability of VAEs for robust, unsupervised quantification of extremes even when underlying anomaly distributions do not strictly conform to Gaussian or symmetric distributions.

\subsection{Limitations}
The findings in section~\ref{sec:dis_posneg} reinforce that while methodological differences exist, both machine-learning and traditional spectral approaches can yield convergent estimates of biogeochemical system vulnerability, provided their limitations and biases are carefully interpreted in context.

Limitations of the current approach include the relatively simple VAE architecture and limited exploration of alternative latent space dimensions. Future work should investigate more sophisticated temporal modeling approaches, such as recurrent VAE architectures with Long Short Term Memory (LSTM) layers, and explore the interpretability of latent space representations for understanding physical drivers of extreme events.
\section{Conclusion}

This study successfully demonstrates the application of variational autoencoders for detecting extreme events in terrestrial carbon cycle dynamics using CESM2 model output. The VAE approach shows strong agreement with established SSA methods in identifying spatial and temporal patterns of carbon cycle extremes, while offering computational advantages and enhanced capability for non-linear pattern recognition. While SSA methods require the temporal periodicity of trends and seasonal cycle in the data to be defined based on expert knowledge, VAE inherently discovers them from the data.

Key findings include: (1) VAE methods consistently produce higher threshold values than SSA but maintain similar spatial and temporal pattern identification; (2) both methods project intensification of extreme events toward 2050--80, with the CNA region showing the most severe increases; (3) spatial distributions of extreme event frequencies show remarkable consistency between methods, supporting the robustness of identified hotspot regions; and (4) the temporal patterns of extreme events reveals concerning trends toward more frequent and severe carbon cycle disruptions under projected future conditions.

The successful application of VAE methods opens opportunities for enhanced detection of extreme events using deep learning approaches. Future research should explore more sophisticated architectures and investigate the physical interpretability of learned representations. The methodology presented provides a foundation for operational extreme event monitoring and improved understanding of carbon-climate feedbacks under changing environmental conditions.
The increasing frequency and magnitude of projected extreme events across all regions underscore the urgent need for enhanced monitoring and prediction capabilities. The convergence of machine learning and traditional analytical approaches demonstrated here offers promising pathways for advancing our understanding and early warning capabilities.

\ifanon
\else
\section*{Acknowledgment}

This research was supported by the Reducing Uncertainties in Biogeochemical Interactions through Synthesis and Computation (RUBISCO) Science Focus Area (RUBISCO SFA KP1703), which is sponsored by the Regional and Global Model Analysis (RGMA) activity of the Earth \& Environmental Systems Modeling (EESM) Program in the Earth and Environmental Systems Sciences Division (EESSD) of the Office of Biological and Environmental Research (BER) in the US Department of Energy Office of Science. This manuscript has been authored by UT-Battelle, LLC, under contract DE-AC05-00OR22725 with the US Department of Energy (DOE). The US government retains and the publisher, by accepting the article for publication, acknowledges that the US government retains a nonexclusive, paid-up, irrevocable, worldwide license to publish or reproduce the published form of this manuscript, or allow others to do so, for US government purposes. DOE will provide public access to these results of federally sponsored research in accordance with the DOE Public Access Plan (\url{http://energy.gov/downloads/doe-public-access-plan}).

This research used resources of the National Energy Research Scientific Computing Center (NERSC), a U.S. Department of Energy Office of Science User Facility supported by the Office of Science of the U.S. Department of Energy under Contract No. DE-AC02-05CH11231.

\fi

\section*{Open Research}
\label{sec: openresearch}
\ifanon
\else

The CESM2 GPP data in this study are available from the Earth System Grid Federation (ESGF)(\url{https://doi.org/10.22033/ESGF/CMIP6.7768}, \cite{esgf_data}). 
Codes for data preprocessing and computing SSA-based anomalies and extremes are available at  \url{https://doi.org/10.5281/zenodo.7854623}, \cite{Sharma_Codes_Zenodo}.
Codes for computing anomalies and extremes using VAE are available at \url{https://doi.org/10.5281/zenodo.17187645}, \cite{sharma_DMESS2025_zenodo}.
\fi

\bibliographystyle{IEEEtran}
\bibliography{references}

\end{document}